\documentclass[10pt,twocolumn,letterpaper]{article}

\usepackage{cvpr}
\usepackage{times}
\usepackage{epsfig}
\usepackage{graphicx}
\usepackage{amsmath}
\usepackage{amssymb}

\usepackage{helvet}
\usepackage{courier}
\usepackage{amsthm}
\usepackage{amsfonts}
\usepackage{algorithm}
\usepackage{algpseudocode}
\usepackage{multirow}
\usepackage{booktabs}

\newcommand{\bs}{\boldsymbol}
\algnewcommand\algorithmicinput{\textbf{Input:}}
\algnewcommand\INPUT{\item[\algorithmicinput]}
\algnewcommand\algorithmicoutput{\textbf{Output:}}
\algnewcommand\OUTPUT{\item[\algorithmicoutput]}
\usepackage{color}

\usepackage{soul}

\newcommand{\mb}{\mathbf}

\usepackage[pagebackref=true,breaklinks=true,letterpaper=true,colorlinks,bookmarks=false]{hyperref}

\usepackage{url}
\usepackage{relsize}
\usepackage[svgnames]{xcolor}
\hypersetup{
  colorlinks,
  urlcolor=Blue}

\cvprfinalcopy % *** Uncomment this line for the final submission

\def\cvprPaperID{2551} % *** Enter the CVPR Paper ID here

\ifcvprfinal\fi
\begin{document}

%%%%%%%%% TITLE
\title{Understanding Traffic Density from Large-Scale Web Camera Data} 
%\footnote{Accepted by IEEE Conference on Computer Vision and Pattern Recognition (CVPR) 2017}

\iffalse
\author{Shanghang Zhang\\
Carnegie Mellon University\\
5000 Forbes Ave. Pittsburgh\\
{\tt\small shanghaz@andrew.cmu.edu}
\and
Guanhang Wu\\
Carnegie Mellon University\\
5000 Forbes Ave. Pittsburgh\\
{\tt\small guanhanw@andrew.cmu.edu}
\and

Joao Costeira\\
Institution2\\
First line of institution2 address\\
{\tt\small secondauthor@i2.org}

\and
Second Author\\
Institution2\\
First line of institution2 address\\
{\tt\small secondauthor@i2.org}

}
\fi

\iffalse
\author{Shanghang~Zhang\thanks{S. Zhang is with ECE, Carnegie Mellon University, Pittsburgh, PA, 15213 USA, shanghaz@andrew.cmu.edu. This research was partially supported by Fundação para a Ciência e a Tecnologia (SFRH/BD/113729/2015) and a PhD grant from the Carnegie Mellon-Portugal program},~
        Guanhang~Wu\thanks{G. Wu is with CS, CMU, Pittsburgh, guanhanw@andrew.cmu.edu.},~
        Jo$\tilde{a}$o~Paulo~Costeira\thanks{J. Costeira is with the ECE, Instituto Superior T$\acute{e}$cnico, Lisbon, Portugal, jpc@isr.ist.utl.pt},~
        and~Jos$\acute{e}$~M.~F.~Moura\thanks{J. Moura is with ECE, CMU, Pittsburgh, moura@andrew.cmu.edu.}% <-this % stops a space
}
\fi

\author{Shanghang Zhang${^\dagger}{^,}{^\ddagger}$, Guanhang Wu$^\dagger$, Jo\~{a}o P. Costeira$^\ddagger$, Jos\'{e} M. F. Moura$^\dagger$\\$^\dagger$Carnegie Mellon University, Pittsburgh, PA, USA\\$^\ddagger$ISR - IST, Universidade de Lisboa, Lisboa, Portugal\\  {\small\texttt{\{shanghaz, guanhanw\}@andrew.cmu.edu, jpc@isr.ist.utl.pt, moura@andrew.cmu.edu}}}

\maketitle
%\thispagestyle{empty}

%%%%%%%%% ABSTRACT
\begin{abstract}
Understanding traffic density from large-scale web camera (webcam) videos is a challenging problem because such videos have low spatial and temporal resolution, high occlusion and large perspective. To deeply understand traffic density, we explore both optimization based and deep learning based methods. To avoid individual vehicle detection or tracking, both methods map the dense image feature into vehicle density, one based on rank constrained regression and the other based on fully convolutional networks (FCN). The regression based method learns different weights for different blocks of the image to embed road geometry and significantly reduce the error induced by camera perspective. The FCN based method jointly estimates vehicle density and vehicle count with a residual learning framework to perform end-to-end dense prediction, allowing arbitrary image resolution, and adapting to different vehicle scales and perspectives. We analyze and compare both methods, and get insights from optimization based method to improve deep model. Since existing datasets do not cover all the challenges in our work, we collected and labelled a large-scale traffic video dataset, containing 60 million frames from 212 webcams. Both methods are extensively evaluated and compared on different counting tasks and datasets. FCN based method significantly reduces the mean absolute error (MAE) from 10.99 to 5.31 on the public dataset TRANCOS compared with the state-of-the-art baseline.
\end{abstract}

%%%%%%%%% BODY TEXT
\vspace{-0.4cm}
\section{Introduction}

\begin{figure}[t]
\setlength{\abovecaptionskip}{-1cm}
\setlength{\belowcaptionskip}{-20pt}
\begin{center}
\includegraphics[width=0.9\columnwidth, height = 3.9cm]{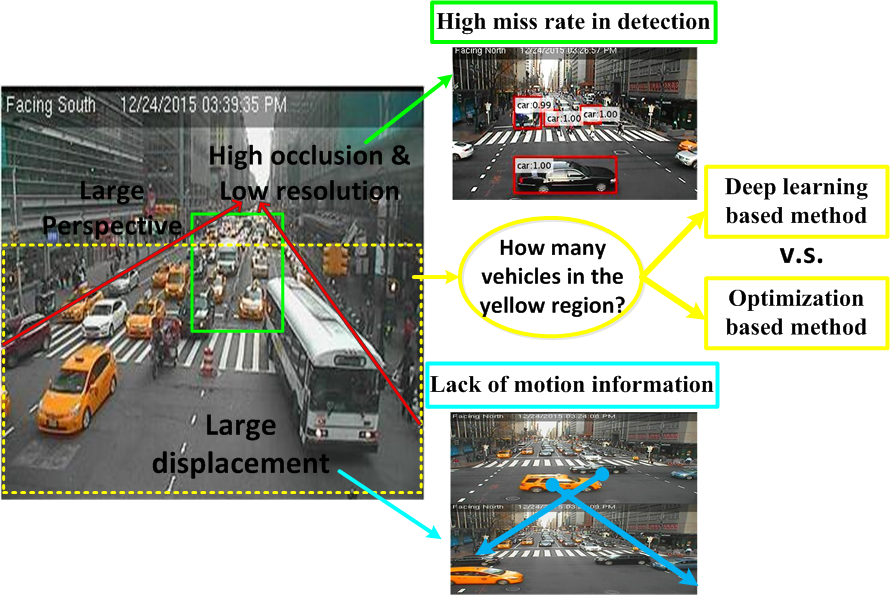}
\end{center}
\vspace{-0.2cm}
\caption{Problem Statement}
\label{fig:fig1}
\end{figure}

Traffic congestion leads to the need for a deep understanding of traffic density, which together with average vehicle speed, form the major building blocks of traffic flow analysis \cite{shirani2009store}. Traffic density is the number of vehicles per unit length of a road (e.g., vehicles per km) \cite{kerner2009introduction}. This paper focuses on traffic density estimation from webcam videos, which are of low resolution, low frame rate, high occlusion and large perspective. As illustrated in Figure \ref{fig:fig1}, we select a region of interest (yellow dotted rectangle) in a video stream, and count the number of vehicles in the region for each frame. Then the traffic density is calculated by dividing that number by the region length.

% \vspace{-5cm}
\setlength{\textfloatsep} {0pt plus 2pt minus 20pt}

Nowadays, many cities are being instrumented with surveillance cameras. However, due to network bandwidth limitations, lack of persistent storage, and privacy concerns \cite{du2013network}, these videos present several challenges for analysis (illustrated in Figure \ref{fig:fig1}): (i)  \textit{Low frame rate}. The time interval between two successive frames of a webcam video typically ranges from $1s$ to $3s$, resulting in large vehicle displacement. (ii)  \textit{Low resolution}. The resolution of webcam videos is $352\times240$. Vehicle at the top of a frame can be as small as $5\times5$ pixels. Image compression also induces artifacts. (iii)  \textit{High occlusion}. Cameras installed at urban intersections often capture videos with high traffic congestion, especially during rush hours. (iv)  \textit{Large perspective}. Cameras are installed at high points to capture more video content, resulting in videos with large perspective. Vehicle scales vary dramatically based on their distance to the camera. These challenges make the existing work for traffic density estimation has many limitations.

%-------------------------------------------------------------------------
\vspace{-0.3cm}
\subsection{Limitation of Related Works}

Several traffic density estimation techniques have been developed in the literature, but they perform less accurately on the webcam data due to the above challenges: \\
% \begin{itemize}
\textbf{Detection based methods.} These methods \cite{zheng2012model,Evgeny2015Traffic} try to identify and localize vehicles in each frame. They perform poorly in low resolution and high occlusion videos. Figure \ref{fig:fig1} shows detection results by Faster RCNN \cite{renNIPS15fasterrcnn}. Even though trained on our collected and annotated webcam dataset, it still exhibits very high missing rate. \\
\textbf{Motion based methods.} Several methods \cite{chen2011real,chen2012vehicle,mo2010vehicles} utilize vehicle tracking to estimate traffic flow. These methods tend to fail due to low frame rate and lack of motion information. Figure \ref{fig:fig1} shows a large displacement of a vehicle (black car) in successive frames due to the low frame rate. Some vehicles only appear once in the video and their trajectory cannot be well estimated.\\
\textbf{Holistic approaches.} These techniques \cite{xiablock} perform analysis on the whole image, thereby avoiding segmentation of each object. \cite{gonccalves2012spatiotemporal} uses a dynamic texture model based on Spatiotemporal Gabor Filters for classifying traffic videos into different congestion types, but it does not provide accurate quantitative vehicle densities. \cite{Lempitsky2010learning} formulates the object density as a linear transformation of each pixel feature, with a uniform weight over the whole image. It suffers from low accuracy when the camera has large perspective.\\
\textbf{Deep learning based methods.} Recently, several deep learning based methods have been developed for object counting\cite{zhang2015cross, zhang2016single, onoro2016towards, zhao2016crossing, arteta2016counting}. The network in \cite{zhang2015cross} outputs a 1D feature vector and fit a ridge regressor to perform the final density estimation, which cannot perform pixel-wise prediction and lose the spatial information. The estimated density map cannot have the same size as the input image. \cite{onoro2016towards} is based on fully convolutional networks but the output density map is still much smaller than the input image, because it does not have deconvolutional or upsampling layers. \cite{arteta2016counting} jointly learns density map and foreground mask for object counting, while it does not solve the large perspective and object scale variation problems.
% \end{itemize}

To summarize, detection and motion based methods tend to fail in high congestion, low resolution, and low frame rate videos, because they are sensitive to video quality and environment conditions. The holistic approaches perform poorly in videos with large perspective and variable vehicle scales. Besides, most of existing methods are incapable of estimating the exact number of vehicles. \cite{garg2016real,hua2012real,choe2010traffic,yu2002highway}.

\iffalse
\begin{figure}[ht]
\setlength{\abovecaptionskip}{0.cm}
\setlength{\belowcaptionskip}{-0.1cm}
\begin{center}
\includegraphics[width=0.8\columnwidth, height = 2.5cm]{figs/fig9}
\end{center}
\caption{Failure cases of existing methods. (a) Tracking-based methods are precluded due to large vehicle displacement. (b) High missing rate from Faster RCNN detection results.}
\label{fig:fig9}
\end{figure}
\fi

\vspace{-0.1cm}
\subsection{Contributions}
\label{sec:Contributions}
To deeply understand traffic density and overcome the challenges from real-world webcam data, we explore both deep learning based and optimization based methods. The optimization based model (OPT-RC) embeds scene geometry through the rank constraint of multiple block-regressors, and motivates the deep-learning model FCN-MT. FCN-MT shares the idea of mapping local feature into vehicle density with OPT-RC, while it replaces the BG subtraction, feature extractor, and block-regressors with fully convolutional networks. With extensive experiments, we analyze and compare both methods, and get insights from optimization based method to improve deep model. 

\begin{figure}[t]
\begin{center}
\includegraphics[width=0.95\columnwidth, height = 3.2cm]{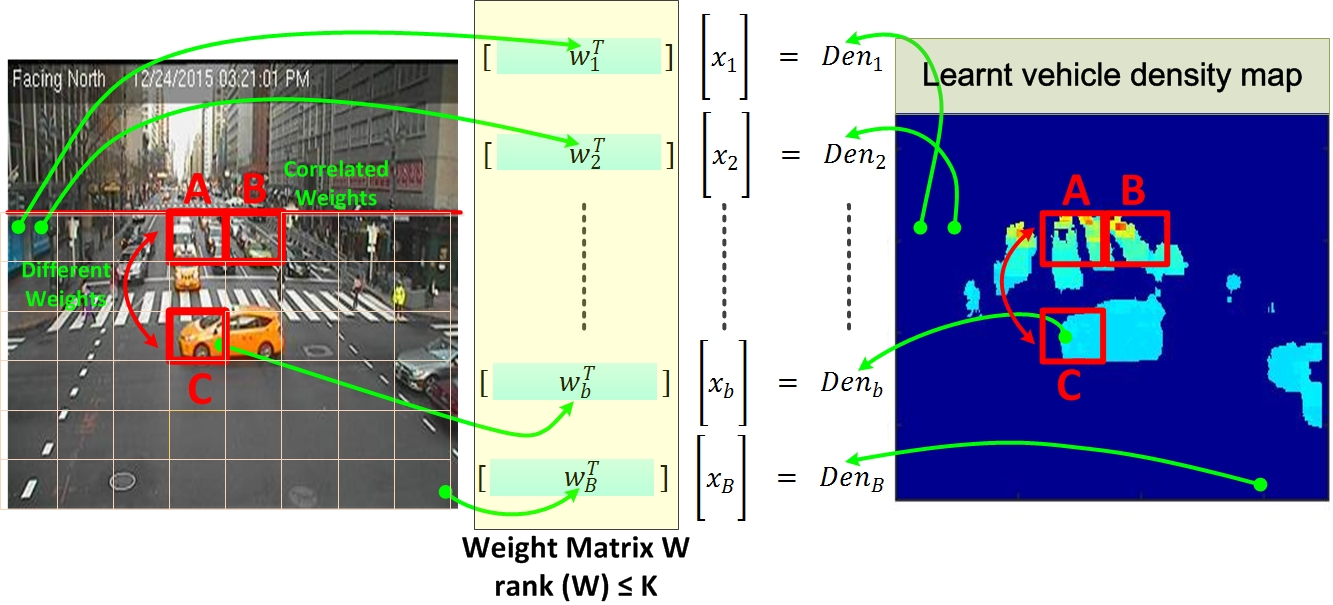}
\end{center}
\vspace{-0.4cm}
\caption{Intuition for OPT-RC.}
\vspace{0.2cm}
\label{fig:fig2}
\end{figure}

%\setlength{\textfloatsep} {0pt plus 2pt minus 20pt}

% \begin{itemize}
\textbf{Optimization Based Vehicle Density Estimation with Rank Constraint (OPT-RC)}.
Inspired by \cite{Lempitsky2010learning}, which maps each pixel feature into vehicle density with a uniform weight, we propose a regression model to learn different weights for different blocks to increase the degrees of freedom on the weights, and embed geometry information. It outperforms work \cite{Lempitsky2010learning} and obtains high accuracy in low quality webcam videos, especially overcoming large perspective challenge. We first divide the target region into blocks, extract features for each block, and subtract background. As illustrated in Figure \ref{fig:fig2}, we linearly map each block feature $\mb{x}_b$ into vehicle density $\textrm{Den}_b=\mb{w}_b^\top \mb{x}_b$. To avoid large errors induced by large perspective, we build one regressor per block with different weights $\mb{w}_b$, and learn the optimal weights. All the weight vectors are stacked into a weight matrix $\mb{W} = \begin{bmatrix} \mb{w}_{1}^\top; &\mb{w}_{2}^\top;  &...  &\mb{w}_{B}^\top \end{bmatrix}$. To handle high dimensionality and capture the correlations among weight vectors of different blocks, rank constraint is imposed on the weight matrix $\mb{W}$. The motivation behind such treatment is illustrated in Figure \ref{fig:fig2}. Due to large perspective, vehicles have different scales in block A and C, and their corresponding vehicle densities are also different. Yet for block A and B, the vehicle densities are similar. Thus we build the weight matrix $\mb{W}$ to reflect both the diversity and the correlation among weight vectors. After the vehicle density map is estimated, the vehicle count can be obtained by integrating the vehicle density map. Finally, the traffic density is obtained by dividing the vehicle counts by the length of the target region.

\begin{figure}[ht]
\setlength{\abovecaptionskip}{-0.9cm}
\setlength{\belowcaptionskip}{-0.9cm}
\begin{center}
\includegraphics[width= \columnwidth, height = 3.5cm]{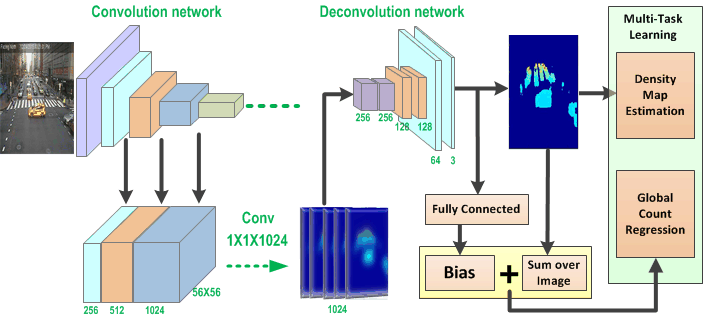}
\end{center}
\vspace{-0.3cm}
\caption{Framework of FCN-MT}
\vspace{-0.3cm}
\label{fig:fig8}
\end{figure}
\setlength{\textfloatsep} {0pt plus 2pt minus 20pt}

% \end{itemize}

\iffalse
\begin{figure}
\setlength{\abovecaptionskip}{0.cm}
\setlength{\belowcaptionskip}{-0.5cm}
\begin{center}
\includegraphics[width=0.8\columnwidth, height = 1.8cm]{figs/fig10.pdf}
\end{center}
\caption{Vehicle density map from perspective. Due to camera perspective, block A and C have different vehicle densities, while block A and B has similar vehicle densities.}
\label{fig:fig10}
\end{figure}
\fi

\textbf{FCN Based Multi-Task Learning for Vehicle Counting (FCN-MT)}. To avoid individual vehicle detection or tracking, besides the proposed optimization based model, we further propose an FCN based model to jointly learn vehicle density and vehicle count. The framework is illustrated in Figure \ref{fig:fig8}. To produce the density map that has the same size as input image, we design a fully convolutional network \cite{long2015fully} to perform pixel-wise prediction whole-image-at-a-time by dense feedforward computation and backpropagation. Instead of applying simple bilinear interpolation for upsampling, we add deconvolution layers on top of convolutional layers, whose parameters can be learnt during training.

There are two challenges of FCN-based object counting: (1) object scale variation, and (2) reduced feature resolution\cite{chen2016deeplab}. To avoid large errors induced by scale variations, we jointly perform global count regression and density estimation. Single task (density estimation) method only encourages networks to approximate ground truth density and directly sums the densities to get the count, which suffers from large error when there is extreme occlusion or oversized vehicles. Yet the multi-task framework is fundamental to account for such deviations, enabling related objectives achieve better local optima, improving robustness, and providing more supervised information. Furthermore, instead of directly regressing the global vehicle count from the last feature map, we develop a residual learning framework to reformulate global count regression as learning residual functions with reference to the sum of densities in each frame. Such design avoids learning unreferenced functions and eases the training of network. The second challenge is caused by the repeated combination of max-pooling and striding. To solve this problem, we produce denser feature maps by combining appearance features from shallow layers with semantic features from deep layers. We then add a convolution layer after the combined feature volume with 1x1 kernels to perform feature re-weighting. The re-weighted features better distinguish foreground and background. Thus the whole network is able to accurately estimate vehicle density without foreground segmentation.

%for the DCNNs are originally designed for image classification\cite{krizhevsky2012imagenet},\cite{simonyan2014very},\cite{szegedy2015going} This results in feature maps with significantly reduced spatial resolution.

\textbf{Webcam Traffic Video Dataset (WebCamT)} We collected and labelled a large-scale webcam traffic dataset, which contains 60 million frames from 212 webcams installed in key intersections of the city. This dataset is annotated with vehicle bounding box, orientation, re-identification, speed, category, traffic flow direction; weather and time. Unlike existing car dataset KITTI\cite{Geiger2012CVPR} and Detrac\cite{DETRAC:CoRR:WenDCLCQLYL15}, which focus on vehicle models, our dataset emphasizes real world traffic network analysis in a large metropolis. This dataset has three benefits: (i) It motivates research on vision based traffic flow analysis, posing new challenges for state-of-the-art algorithms. (ii) With various street scenes, it can serve as benchmark for transfer learning and domain adaptation. (iii) With large amount of labeled data, it provides training set for various learning based models, especially for deep learning based techniques.

Contributions of this paper are summarized as follows:

1. We propose an optimization based density estimation method (OPT-RC) that embeds road geometry in the weight matrix and significantly reduces error induced by perspective. It avoids detecting or tracking individual vehicles.

2.We propose FCN based multi-task learning to jointly estimate vehicle density and count with end-to-end dense prediction. It allows arbitrary input image resolution, and adapts to different vehicle scale and perspective. 
% It avoids background subtraction by performing feature selection to increase the discrimination of features.

3. We collect and annotate a large-scale webcam traffic dataset, which poses new challenges to state-of-the-art traffic density estimation algorithms. To the best of our knowledge, it is the first and largest webcam traffic dataset with elaborate annotations.

4. With extensive experiments on different counting tasks, we verify and compare the proposed FCN-MT with OPT-RC, and obtain insight for future study.

The rest of paper is outlined as follows. Section 2 introduces the proposed OPT-RC. Section 3 introduces the proposed FCN-MT. Section 4 presents experimental results, and Section 5 compares OPT-RC with FCN-MT.

%------------------------------------------------------------------------
\vspace{-0.2cm}
\section{Optimization Based Vehicle Density Estimation with Rank Constraint}

To overcome limitations of existing work, we propose a block-level regression model with rank constraint. The overall framework is described in Section \ref{sec:Contributions}. We first perform foreground segmentation based on GrabCut \cite{Rother2004Grabcut}. To automate the segmentation process, we initialize the background and foreground based on the difference between the input frame and the pure background image, which is generated by using a reference image with no vehicles taken in light traffic periods and transferring it to other time periods by brightness adjustment. We assume that a stream of $N$ images $I_1,\cdots,I_N$ are given, for which we select a region of interest and divide it into $J$ blocks. The block size can vary from $16\times 16$ to $1\times 1$, depending on the width of the lane and the length of the smallest vehicle. A block $B_j^{(i)}$ in each image $I_i$ is represented by a feature vector $\mb{x}^{(i)}_j\in\mathbb{R}^K$. Examples of particular choices of features are given in the experimental section. It is assumed that each training image $I_i$ is annotated with a set of 2D bounding boxes, centered at pixels $P_i=\{p_1,\cdots,p_{c(i)}\}$, where $c(i)$ is the total number of annotated vehicles in the $i$-th image. The density functions in our approach are real-valued functions over pixel grids, whose integrals over image regions equal to the vehicle counts. For a training image $I_i$, we calculate the ground truth density based on the labeled bounding boxes (shown in Figure \ref{fig:fig5}). The pixel $p$ covered by a set of bounding boxes $O(p)$ has density $D(p)$ defined as
\begin{equation}
D(p)=\sum_{o\in O(p)}\frac{1}{A(o)},
\end{equation}
where $A(o)$ denotes the area of bounding box $o$. Then, we define the density $D(B_j)$ of a block as
\begin{equation}
D(B_j)=\sum_{p\in B_j}D(p).
\end{equation}
Given a set of training images together with their ground truth densities, for each block $B_j$, we learn a block-specific linear regression model that predicts block-level density $\widehat{D}(B_j)$ given its feature representation $\mb{x}_j$ by
\begin{equation}
\label{eq:predict}
\widehat{D}(B_j)=\mb{w}_j^\top \mb{x}_j,
\end{equation}
where $\mb{w}_j\in \mathbb{R}^K$ is the coefficient vector of the linear regression model to be learned for block $j$. We assign different weights to different blocks. To capture the correlations and commonalities among the regression weight vectors at different blocks, we encourage these vectors to share a low-rank structure. To avoid overfitting, we add $\ell_2$-regularization to these weight vectors. To encourage sparsity of weights, $\ell_1$-regularization is imposed. Let $\mb{W}\in\mathbb{R}^{K\times J}$, where the $j$-th column vector $\mb{w}_j$ denotes the regression coefficient vector of block $j$. To this end, we define the following regularized linear regression model with low-rank constraint
\begin{equation}
\begin{array}{lll}
\underset{\mb{W}}{\text{min}} \!\!\!& \frac{1}{2N}\sum\limits_{i=1}^{N}\sum\limits_{j=1}^{J}(\mb{w}_j^\top \mb{x}^{(i)}_j-D(B_j^{(i)}))^2+\alpha\|\mb{W}\|_F^2+\beta|W|_1\\
\textrm{s.t.} & \text{rank}(\mb{W})\leq r
\end{array}
\end{equation}
%This problem is challenging to solve. First, the objective function contains an $\ell_1$ regularizer, which is non-smooth. Second, the parameter matrix $\mb{W}$ is constrained to be of low-rank.
%To cope with these two challenges,
To solve this rank-constrained problem that has a non-smooth objective function, we develop an accelerated projected subgradient descent (APSD) \cite{li2015accelerated} algorithm outlined in Algorithm \ref{alg:pgd}, which iteratively performs subgradient descent, rank projection, and acceleration. Two sequences of variables $\{\mb{A}_k\}$ and $\{\mb{W}_k\}$ are maintained for the purpose of performing acceleration. \\
% \begin{itemize}
\textbf{Subgradient Descent}
Subgradient descent is performed over variable $\mb{A}_k$. We first compute the subgradient $\bigtriangleup \mb{A}_k$ of the non-smooth objective function $\frac{1}{2N}\sum\limits_{i=1}^{N}\sum\limits_{j=1}^{J}(\mb{a}_j^\top \mb{x}^{(i)}_j-D(B_j^{(i)}))^2+\alpha\|\mb{A}\|_F^2+\beta|\mb{A}|_1$, where the first and second terms are smooth, hence their subgradients are simply gradients. For the third term $|\mb{A}|_1$ which is non-smooth, its subgradient $\partial \mb{A}$ can be computed as
\begin{equation}
\partial A_{ij}=
\begin{cases}
    +1      & \quad \text{if } A_{ij}\geq 0\\
    -1  & \quad \text{if } A_{ij}<0\\
  \end{cases}
\end{equation}
Adding the subgradients of the three terms, we obtain the subgradient $\bigtriangleup \mb{A}_k$ of the overall objective function. Then $\mb{A}_k$ is updated by
\begin{equation}
\mb{A}_k\gets \mb{A}_k-t_k \bigtriangleup \mb{A}_k.
\end{equation}
\textbf{Rank Projection}
In lines 4-5, we project the newly obtained $\mb{A}_k$ to the feasible set $\{\mb{W}|\text{rank}(\mb{W})\leq r\}$, which amounts to solving the following problem

\begin{equation}
\begin{array}{ll}
\text{min}_{\mb{W}} & \|\mb{A}_k-\mb{W}\|_{F}^2 \\
\text{s.t.}     & \text{rank}(\mb{W})\leq r
\end{array}
\end{equation}

According to \cite{eckart1936approximation}, the optimal solution $\mb{W}^*$ can be obtained by first computing the largest $r$ singular values and singular vectors of $\mb{A}_k$: $\mb{U}_r$, $\bs\Sigma_r$, $\mb{V}_r$, then setting $\mb{W}^*$ to $\mb{U}_r\bs\Sigma_r\mb{V}_r^\top$. \\
\textbf{Acceleration}
In lines 6-7, acceleration is performed by updating the step size according to the following rule: $t_{k+1}\gets \frac{1}{2}(1+\sqrt{1+4t_k^2})$, and adding a scaled difference between consecutive $\mb{W}$s to $\mb{A}$.
% \end{itemize}

It is worth noting that this optimization problem is not convex and the APSD algorithm may lead to local optimal. It helps to run the algorithm multiple times with different random initializations.

\begin{algorithm}[t]
\caption{Accelerated Projected Subgradient Descent}\label{alg:pgd}
\begin{algorithmic}[1]
\INPUT Data $\mathcal{D}$, rank $r$, regularization parameters $\alpha$, $\beta$
\While{not converged}
\State Compute gradient $\bigtriangleup \mb{A}_k$
\State $\mb{A}_k\gets \mb{A}_k-t_k \bigtriangleup \mb{A}_k$
\State Compute top $r$ singular values and vectors of $\mb{A}_k$:
 $\mb{U}_r$, $\bs\Sigma_r$, $\mb{V}_r$
\State $\mb{W}_{k+1}\gets \mb{U}_r\bs\Sigma_r\mb{V}_r^\top$
\State $t_{k+1}\gets \frac{1}{2}(1+\sqrt{1+4t_k^2})$
\State $\mb{A}_{k+1}\gets \mb{A}_k\frac{t_{k-1}-1}{t_k}(\mb{W}_k-\mb{W}_{k-1})$
\EndWhile
\OUTPUT $\mb{W}\gets \mb{W}_{k+1}$
\end{algorithmic}
\end{algorithm}

\setlength{\textfloatsep} {0pt plus 2pt minus 20pt}
%-------------------------------------------------------------------------

\vspace{-0.2cm}
\section{FCN Based Multi-Task Learning}
We also propose an FCN based model to jointly learn vehicle density and global count. The vehicle density estimation can be formulated as $D(i)=F(X_{i}; \Theta )$, where $X_{i}$ is the input image, $\Theta$ is the set of parameters of the FCN-MT model, and $D(i)$ is the estimated vehicle density map for image i. The ground truth density map can be generated in the same way as Section 2.
\vspace{-0.2cm}
\subsection{Network Architecture}
\label{sec:3.1}
Inspired by the FCN used in semantic segmentation\cite{long2015fully}, we develop FCN to estimate the vehicle density. After the vehicle density map is estimated, the vehicle count can be obtained by integrating the vehicle density map. However, we observed that variation of vehicle scales induces error during the direct integration. Especially, the large buses/trucks (oversized vehicles) in close view induce sporadically large errors in the counting results. To solve this problem, we propose a deep multi-task learning framework based on FCN to jointly learn vehicle density map and vehicle count. Instead of directly regressing the count from the last feature map or the learnt density map, we develop a residual learning framework to reformulate global count regression as learning residual functions with reference to the sum of densities in each frame. The overall structure of our proposed FCN-MT is illustrated in Figure \ref{fig:fig8}, which contains convolution network, deconvolution network, feature combination and selection, and multi-task residual learning.

The convolution network is based on pre-trained ResNets\cite{he2016deepnew}. The pixel-wise density estimation requires high feature resolution, yet the pooling and striding reduce feature resolution significantly. To solve this problem, we rescale and combine the features from $2a$, $3a$, $4a$ layers of ResNets. We then add a convolution layer after the combined feature volume with 1x1 kernels to perform feature re-weighting. By learning parameters of this layer, the re-weighted features can better distinguish foreground and background pixels. We input the combined feature volume into the deconvolution network, which contains five deconvolution layers. Inspired by the deep VGG-net\cite{simonyan2014very}, we apply small kernels of 3x3 in the deconvolution layers. The feature is mapped back to the image size by deconvolution layers, whose parameters can be learnt from the training process\cite{noh2015learning}. A drawback of deconvolution layer is that it may have uneven overlap when the kernel size is not divisible by the stride. We add one convolution layer with 3x3 kernels to smooth the checkerboard artifacts and alleviate this problem. Then one more convolution layer with 1x1 kernels is added to map the feature maps into density map.
%Accuracy of our approach can be improved by $8\%$ using ResNets compared with using VGG-net\cite{simonyan2014very}.

\vspace{-0.1cm}
\subsection{Multi-Task Learning}
At the last stage of the network, we jointly learn vehicle density and count. The vehicle density is predicted by the last convolution 1x1 layer from the feature map. Euclidean distance is adopted to measure the difference between the estimated density and the ground truth. The loss function for density map estimation is defined as follows:
\vspace{-0.1cm}
\begin{equation}
L_{D}(\Theta)=\frac{1}{2N}\sum_{i = 1}^{N}\sum_{p=1}^{P}  \left \|  F(X_{i}(p)); \Theta ) - F_{i}(p))\right \|_{2}^{2},
\end{equation}
\vspace{-0.1cm}
where $N$ is the number of training images and $F_{i}(p)$ is the ground truth density for pixel $p$ in image $i$.

For the second task, global count regression, we reformulate it as learning residual functions with reference to the sum of densities, which consists of two parts: (i) base count: the integration of the density map over the whole image; (ii) offset count: predicted by two fully connected layers from the feature map after the convolution 3x3 layer of the deconvolution network. We sum the two parts together to get the estimated vehicle count, as shown in the following equation:
\vspace{-0.1cm}
\begin{equation}
C(i)=G(D(i);\gamma ) + \sum_{p=1}^{P}D(i,p),
\end{equation}
\vspace{-0.1cm}
where $\gamma$ is the learnable parameters of the two fully connected layers, $D(i,p)$ indicates the density of each pixel $p$ in image $i$. We hypothesize that it is easier to optimize the residual mapping than to optimize the original, unreferenced mapping. Considering that the vehicle count for some frames may have very large value, we adopt Huber loss to measure the difference between the estimated count and the ground truth count. The count loss for one frame is defined as follows:
\begin{equation}
L_{\delta }(i)=\left\{\begin{matrix}
\frac{1}{2}(C(i)-C_{t}(i))^{2} &   \textrm{for} |C(i)-C_{t}(i)|\leq \delta, \\
 \delta |C(i)-C_{t}(i)|-\frac{1}{2}\delta ^{2}&  \textrm{otherwise.}
\end{matrix}\right.
\end{equation}
where $C_{t}(i)$ is the ground truth vehicle count of frame $i$, $C(i)$ is the estimated loss of frame $i$. $\delta$ is the threshold to control the outlier in the training sets. Then overall loss function for the network is defined as:
\vspace{-0.1cm}
\begin{equation}
L= L_{D} + \lambda \frac{1}{N}\sum_{i=1}^{N}L_{\delta }(i),
\end{equation}
\vspace{-0.1cm}
where $\lambda$ is the weight of count loss, and is tuned to achieve best accuracy. By simultaneously learning the two related tasks, each task can be better trained with much fewer parameters. The loss function is optimized via batch-based Adam and backpropagation. As FCN-MT adapts to different input image resolutions and variation of vehicle scales and perspectives, it is robust to different scenes. 

% To improve training efficiency and guarantee the performance of both tasks, we first train the network with single density estimation task and then add the loss of global count regression to train the two tasks together. If the target domain contain quite different scenes, we can fine tune the network with only a few training images with the convolution network fixed. With such treatment, the knowledge learnt in the source domain can be preserved, and the models can be adapted to the target domain by fine-tuning the last few layers.

\vspace{-0.2cm}
\section{Experiments}
We extensively evaluate the proposed methods on different datasets and counting tasks: (i) We first introduce our collected and annotated webcam traffic dataset (WebCamT).  (ii) Then we evaluate and compare the proposed methods with state-of-the-art methods on WebCamT dataset and present an interesting application to detect the change of traffic density patterns in NYC on Independence Day. (iii) We evaluate our proposed methods on the public dataset TRANCOS\cite{onoro2016towards}. (iv) We evaluate our methods on the public pedestrian counting dataset UCSD \cite{chan2008privacy} to verify the robustness and generalization of our model.
\vspace{-0.1cm}
\subsection{Webcam Traffic Data Collection}

As there is no existing labeled real-world webcam traffic dataset, in order to evaluate our proposed method, we utilize existing traffic web cameras to collect continuous stream of street images and annotate rich information. Different from existing traffic datasets, webcam data are challenging for analysis due to the low frame rate, low resolution, high occlusion, and large perspective. We select 212 representative web cameras, covering different locations, camera perspective, and traffic states.  For each camera, we downloaded videos for four time intervals each day (7am-8am, 12pm-1pm; 3pm-4pm; 6pm-7pm). These cameras have frame rate around 1 frame/second and resolution $352\times240$. Collecting these data for 4 weeks generates 1.4 Terabytes of video data consisting of 60 million frames. To the best of our knowledge, WebCamT is the first and largest annotated webcam traffic dataset to date.

% The statistics of representative cameras are shown in Table \ref{tb:tb2}.
\iffalse
\begin{table}[t]
\small
\centering
\begin{tabular}{@{}|c|c|c|c|@{}}
\toprule
\textbf{Address} & \textbf{Resolution} & \textbf{\begin{tabular}[c]{@{}c@{}}Time \\ Interval\end{tabular}} & \textbf{Description} \\ \midrule
2 Ave @ 42 St & 352*240 & 1.2 & crosstown St. \\ \midrule
8 Ave @ 42 St & 352*240 & 1.3 & crosstown St. \\ \midrule
3 Ave @ 42 St & 352*240 & 1.3 & crosstown St. \\ \midrule
3 Ave @ 49 St & 352*240 & 1.2 & crosstown St. \\ \midrule
FDR @ 79St & 352*240 & 1.1 & Parkway \\ \midrule
FDR@ 23St & 352*240 & 1.2 & Parkway \\ \bottomrule
\end{tabular}
\caption{Statistics for 6 selected webcams}
\label{tb:tb2}
\end{table}
\fi

\iffalse
Figure \ref{fig:fig4} shows the nine selected cameras’ scenes.
\begin{figure}[t]
\begin{center}
\includegraphics[width=0.8\columnwidth, height = 2.5cm]{figs/tabel2}
\end{center}
\caption{Statistics for webcam traffic dataset}
\label{fig:approach_overview}
\end{figure}
\fi

\iffalse
\begin{figure}[t]
\setlength{\abovecaptionskip}{0.cm}
\setlength{\belowcaptionskip}{-0.1cm}
\begin{center}
\includegraphics[width=0.8\columnwidth]{figs/fig4}
\end{center}
\caption{Nine annotated cameras with representative camera position, perspective, and scene content}
\label{fig:fig4}
\end{figure}
\fi

\begin{figure}[t]
\setlength{\abovecaptionskip}{0.cm}
\setlength{\belowcaptionskip}{-0.2cm}
\begin{center}
\includegraphics[width=0.95\columnwidth, height = 3.3cm]{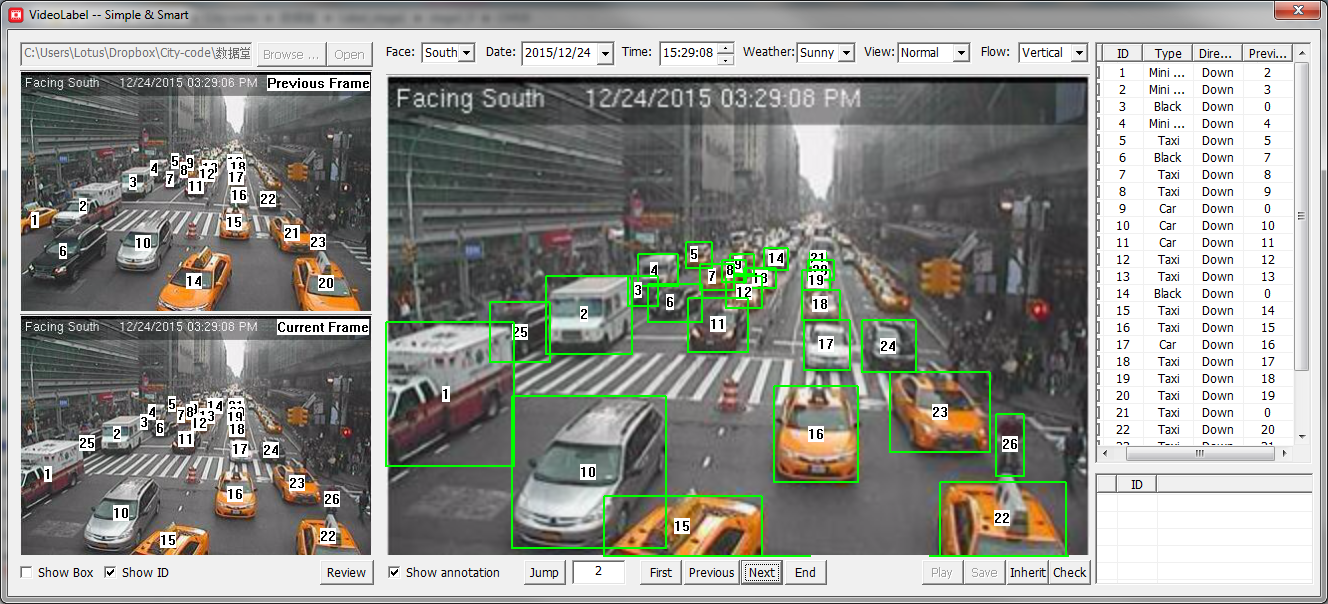}
\end{center}
\vspace{-0.2cm}
\caption{Annotation Instance.}
\label{fig:fig5}
\end{figure}
\setlength{\textfloatsep} {0pt plus 2pt minus 10pt}

We annotate $60,000$ frames with the following information: (i) \emph{Bounding box}: rectangle around each vehicle. (ii) \emph{Vehicle type}: ten types including taxi, black sedan, other cars, little truck, middle truck, big truck, van, middle bus, big bus, other vehicles. (iii) \emph{Orientation}: each vehicle orientation is annotated into four categories: 0, 90, 180, and 270 degrees. (iv) \emph{Vehicle density}: number of vehicles in ROI region of each frame. (v) \emph{Re-identification}: we match the same car in sequential frames. (vi) \emph{Weather}: five types of weather, including sunny, cloudy, rainy, snowy, and intensive sunshine. The annotation for two successive frames is shown in Figure \ref{fig:fig5}. The dataset is divided into training and testing sets, with 45,850 and 14,150 frames, respectively. We select testing videos taken at different time from training videos. WebCamT serves as an appropriate dataset to evaluate our proposed method. It also motivates research on vision based traffic flow analysis, posing new challenges for the state-of-the-art algorithms \footnote{Please email the authors if you are interested in the dataset.}.

\subsection{Quantitative Evaluations on WebCamT}

We evaluate the proposed methods on the testing set of WebCamT, containing 61 video sequences from 8 cameras, and covering different scenes, congestion states, camera perspectives, weather and time of the day. Each video has $352\times240$ resolution, and frame rate around 1 frame/sec. The training set has the same resolution, but is from different videos. Three metrics are employed for evaluation: (i) Mean absolute error (MAE); (ii) Mean square error (MSE); (iii) Average relative accuracy (ARA), which is the average of all the test frame relative errors.

% Details of these 48 videos are described in Table \ref{tb:tb3}.

\iffalse
\begin{table}[]
\centering
\small
\caption{Descriptions of 48 test videos. $C170\sim C511$ are 6 test cameras. Numbers under them are the test frames amount}
\vspace{0.2cm}
\label{tb:tb3}
\resizebox{0.45\textwidth}{!}{%
\begin{tabular}{|l|l|l|l|l|l|l|l|l|}
\hline
Time & Weather & State & C170 & C410 & C495 & C691 & C846 & C511 \\ \hline
4/29/7h & Cloudy & Light & 300 & 550 & 300 & 150 & 300 & 300 \\ \hline
4/29/12h & Cloudy & Medium & 300 & 300 & 300 & 150 & 300 & 300 \\ \hline
4/29/18h & Cloudy & High & 300 & 300 & 300 & 150 & 300 & 300 \\ \hline
5/2/18h & Sunny & High & 300 & 300 & 300 & 300 & 300 & 300 \\ \hline
5/4/18h & Rainy & High & 300 & 550 & 300 & 300 & 0 & 300 \\ \hline
7/4/7h & Sunny & Light & 300 & 300 & 300 & 300 & 300 & 300 \\ \hline
7/4/12h & Sunny & Medium & 300 & 300 & 300 & 300 & 300 & 300 \\ \hline
7/4/18h & Cloudy & High & 300 & 300 & 300 & 300 & 300 & 300 \\ \hline
\hline \multicolumn{3}{l|}{Total Frames:  14150} & 2400 & 2900 & 2400 & 1950 & 2100 & 2400 \\ \hline
\end{tabular}%
}
\end{table}

\setlength{\textfloatsep} {0pt plus 2pt minus 10pt}
\fi

For OPT-RC, to compare with the baseline methods, we extract SIFT feature and learn visual words for each block. The block size is determined by the lane width and the smallest vehicle length. Parameters in Eq.(4) are selected by cross-validation. For FCN-MT, we divide the training data into two groups: downtown and parkway. In each group, we balance the training frames with less than 15 vehicles and the frames with more than 15 vehicles. The network architecture is explained in Section \ref{sec:3.1} and some parameters are shown in Figure \ref{fig:fig8}. The weight $\lambda$ of vehicle count loss in Eq.(11) is 0.1. More details can be found in the released code link: \url{https://github.com/polltooh/traffic_video_analysis}.

\textbf{Baseline approaches}. We compare our method with two methods: \emph{Baseline 1: Learning to count} \cite{Lempitsky2010learning}. It maps the feature of each pixel into object density with uniform weight for the whole image. For comparison, we extract dense SIFT features \cite{lowe2004distinctive} using VLFeat \cite{vedaldi08vlfeat}. The ground truth density is computed as a normalized 2D Gaussian kernel based on the center of each labeled bounding box. \emph{Baseline 2: Hydra}\cite{onoro2016towards}. It learns multi-scale regression networks using a pyramid of image patches extracted at multiple scales to perform final density prediction. We train Hydra 3s model on the same training set as FCN-MT.

\begin{table}[H]
\centering
\caption{Accuracy comparison on WebCamT}
\label{tb:tb4}
\begin{tabular}{|l|l|l|l|l|}
\hline
\multirow{2}{*}{Method} & \multicolumn{2}{c|}{Downtown} & \multicolumn{2}{c|}{Parkway} \\ \cline{2-5}
 & MAE & ARA & MAE & ARA \\ \hline
Baseline 1 & 5.91 &0.5104  &5.19  &0.5248  \\ \hline
Baseline 2 & 3.55 & 0.6613 & 3.64 &0.6741  \\ \hline
OPT-RC &4.56  &0.6102  & 4.24 & 0.6281 \\ \hline
FCN-MT &2.74  &0.7175  & 2.52 &0.784  \\ \hline
\end{tabular}
\end{table}

\setlength{\textfloatsep} {0pt plus 2pt minus 20pt}

\begin{figure}[H]
\setlength{\abovecaptionskip}{0.cm}
\setlength{\belowcaptionskip}{-0.2cm}
\begin{center}
\includegraphics[width=0.9\columnwidth, height = 2.1cm]{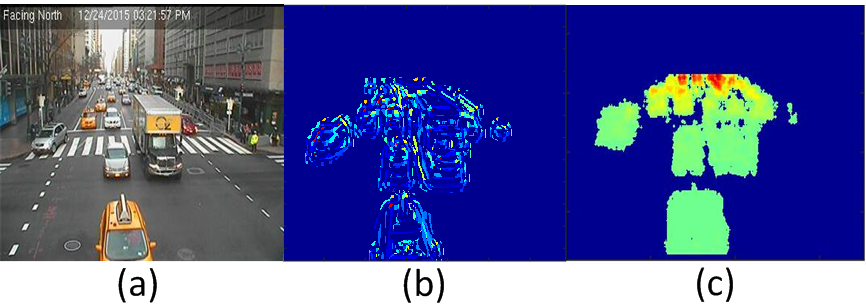}
\end{center}
\vspace{-0.4cm}
\caption{Comparison of OPT-RC and Baseline 1.}
\label{fig:fig14}
\end{figure}

\begin{figure}[H]
\setlength{\abovecaptionskip}{0.cm}
\setlength{\belowcaptionskip}{-0.2cm}
\begin{center}
\includegraphics[width=0.9\columnwidth, height = 6.2cm]{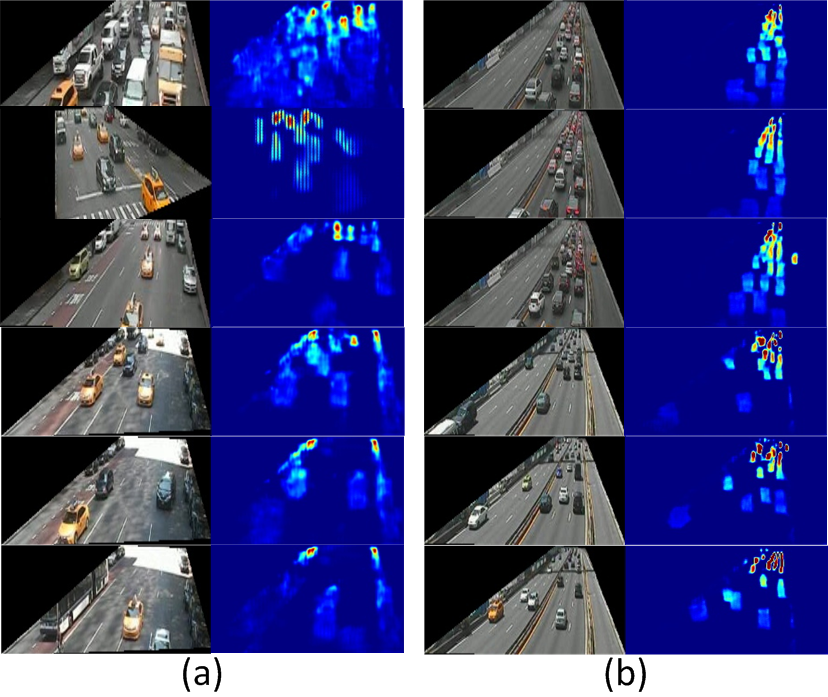}
\end{center}
\vspace{-0.4cm}
\caption{Density map from FCN-MT: (a) Downtown; (b) Parkway. Top three rows are cloudy; Bottom three rows are sunny.}
\label{fig:fig15}
\end{figure}

\textbf{Experimental Results}. Below, we compare the errors of the proposed and baseline approaches in Table \ref{tb:tb4}. From these results, we conclude that FCN-MT outperforms the baseline approaches and OPT-RC for all the measurements. As the testing data cover different congestion states, camera perspectives, weather conditions and time of the day, these results verify the generalization and robustness of FCN-MT. OPT-RC outperforms the non-deep learning based Baseline 1 and shows comparable results with Baseline 2, but require much less training data. Compared with FCN-MT, OPT-RC is less generalizable, but it can learn smooth density map with geometry information. Figure \ref{fig:fig14} shows the original image (a), learned density map from Baseline1 (b), and learned density map from OPT-RC (c). We see that the density map from Baseline 1 can not reflect the perspective present in the video, while density map from our method captures well the camera perspective. Figure \ref{fig:fig15} shows the density map learned from FCN-MT. Without foreground segmentation, the learned density map can still estimate the region of vehicles, and distinguish background from foreground in both sunny and cloudy, dense and sparse scenes. Yet due to the uneven overlaps of the deconvolution layers, checkerboard artifacts are created on the learnt density map.

\begin{figure}[H]
\setlength{\abovecaptionskip}{0.cm}
\setlength{\belowcaptionskip}{-0.5cm}
\begin{center}
\includegraphics[width=0.95\columnwidth, height = 3.7cm]{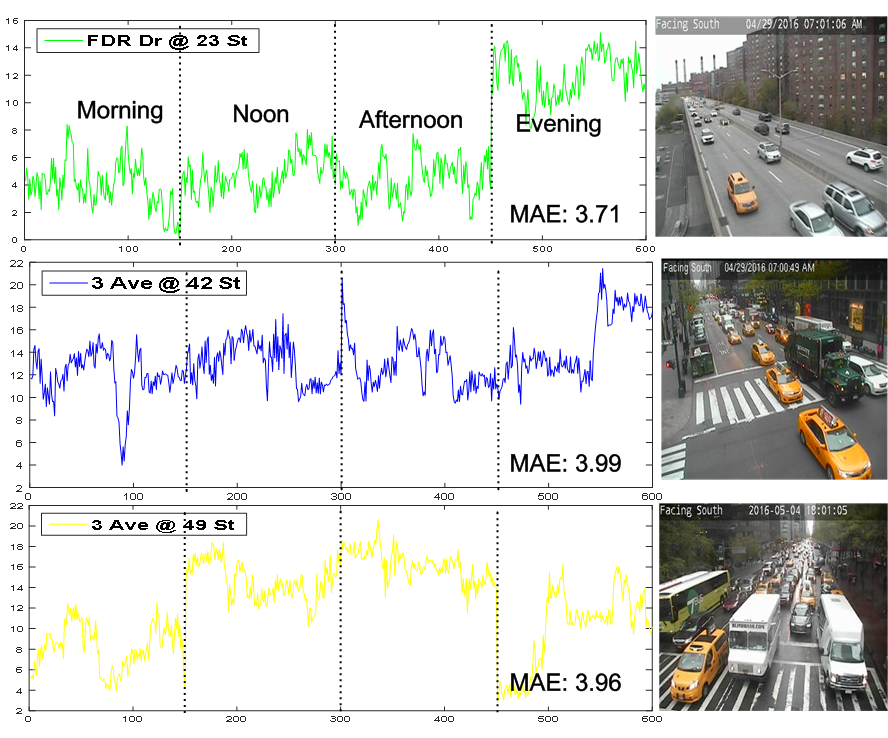}
\end{center}
\vspace{-0.3cm}
\caption{Estimated traffic density from OPT-RC for three cameras. (Left) estimated traffic density curve for each camera, where the X-axis represents frame index and the Y-axis represents traffic density. MAE for each curve is shown at the bottom right of each plot. To show the time series for one day, we select 150 frames for each time interval (morning, noon, afternoon, and evening). (Right) one sample image for each camera.}
\label{fig:OPT}
\end{figure}

\begin{figure}[t]
\setlength{\abovecaptionskip}{0.cm}
\setlength{\belowcaptionskip}{-0.5cm}
\begin{center}
\includegraphics[width=0.95\columnwidth, height = 3.7cm]{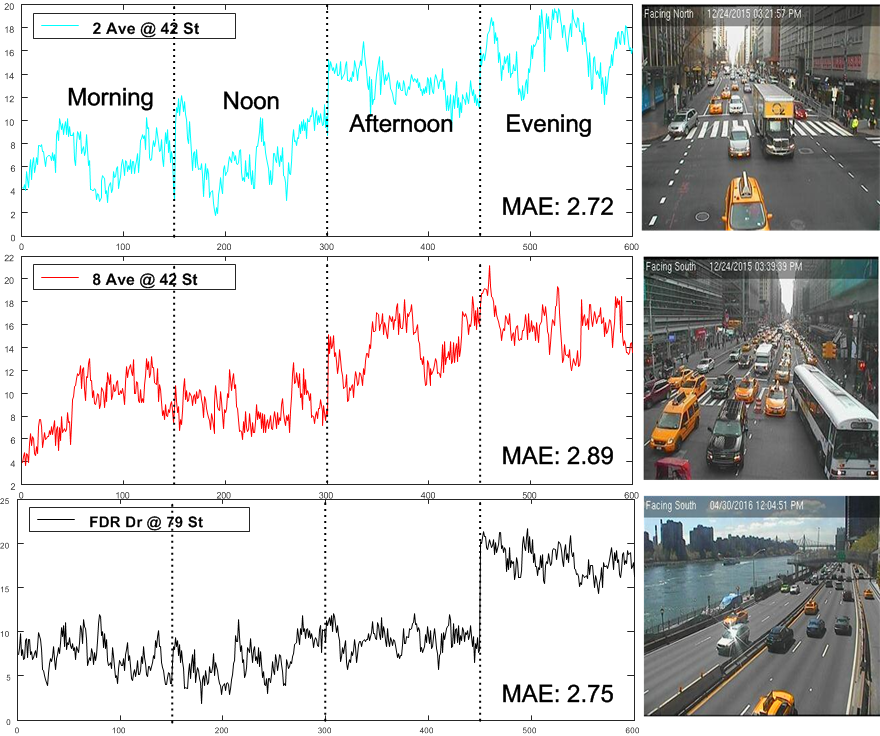}
\end{center}
\vspace{-0.3cm}
\caption{Estimated traffic density from FCN-MT for three cameras. Same setting as Figure 7.}
\label{fig:FCN}
\end{figure}

\vspace{0.2cm}
Figure \ref{fig:OPT} and Figure \ref{fig:FCN} shows the estimated traffic density of OPT-RC and FCN-MT from long time sequences and multiple cameras, respectively. MAE of each camera's estimated traffic density is shown at the bottom right of each plot. From the results, we see that FCN-MT has more accurate estimation than OPT-RC, yet both methods can capture the trends of the traffic density. For the same time of day, traffic density from downtown cameras are on average higher than that from parkway cameras. For the same camera location, traffic density during nightfall (18:00-19:00) is generally higher than at other times of the day. Especially for parkway cameras, nightfall traffic density increases significantly when compared to traffic density in the morning and at noon. As the test videos cover different locations, weather, camera perspectives, and traffic states, those results verify the robustness of the proposed methods.

\setlength{\textfloatsep} {0pt plus 2pt minus 20pt}
\vspace{-0.4cm}

\setlength{\textfloatsep} {0pt plus 2pt minus 20pt}
\vspace{-0.4cm}

\setlength{\textfloatsep} {0pt plus 2pt minus 5pt}
\vspace{-0.4cm}

\setlength{\textfloatsep} {0pt plus 2pt minus 5pt}

\iffalse
\begin{figure}[t]
\setlength{\abovecaptionskip}{0.cm}
\setlength{\belowcaptionskip}{-0.5cm}
\begin{center}
\includegraphics[width=0.95\columnwidth, height = 4.5cm]{figs/fig13.jpg}
\end{center}
\caption{Traffic density without (Top) and with (Bottom) FOV/FRV classification. These two results have the same ground truth traffic density (TrueCount). Dotted circles indicate the frames with oversized vehicles.}
\label{fig:fig13}
\end{figure}
\fi

\iffalse
\begin{table*}[t]
\small
\centering
\begin{tabular}{@{}|c|c|c|c|l|l|l|l|l|l|l|l|l|@{}}
\toprule
\multirow{2}{*}{\textbf{Method}} & \multicolumn{4}{c|}{\textbf{MAE}} & \multicolumn{4}{c|}{\textbf{ARE}} & \multicolumn{4}{c|}{\textbf{MRE}} \\ \cmidrule(l){2-13}
 & V1 & V2 & V3 & V4 & V1 & V2 & V3 & V4 & V1 & V2 & V3 & V4 \\ \midrule
\begin{tabular}[c]{@{}c@{}}Learning to   Count\end{tabular} & 1.95 & 1.65 & 1.93 & 1.55 & 0.15 & 0.13 & 0.14 & 0.13 & 0.11 & 0.11 & 0.11 & 0.10 \\ \midrule
Fastern RCNN & 2.84 & 1.69 & 3.56 & 1.11 & 0.19 & 0.13 & 0.26 & 0.10 & 0.19 & 0.11 & 0.22 & 0.10 \\ \midrule
\begin{tabular}[c]{@{}c@{}}Webcam-targeted   Counting\end{tabular} & 3.75 & 2.47 & 4.15 & 1.63 & 0.25 & 0.19 & 0.30 & 0.14 & 0.26 & 0.18 & 0.26 & 0.12 \\ \midrule
\begin{tabular}[c]{@{}c@{}}Proposed   Method\end{tabular} & 1.57 & 1.25 & 1.54 & 1.22 & 0.12 & 0.10 & 0.12 & 0.10 & 0.10 & 0.06 & 0.10 & 0.05 \\ \bottomrule
\end{tabular}
\caption{Traffic density estimation errors on 4 video clips}
\label{tb:tb4}
\vspace{-0.1in}
\end{table*}
\fi

\begin{figure}[t]
\setlength{\abovecaptionskip}{0.cm}
\setlength{\belowcaptionskip}{-0.5cm}
\begin{center}
\includegraphics[width=0.95\columnwidth, height = 3cm]{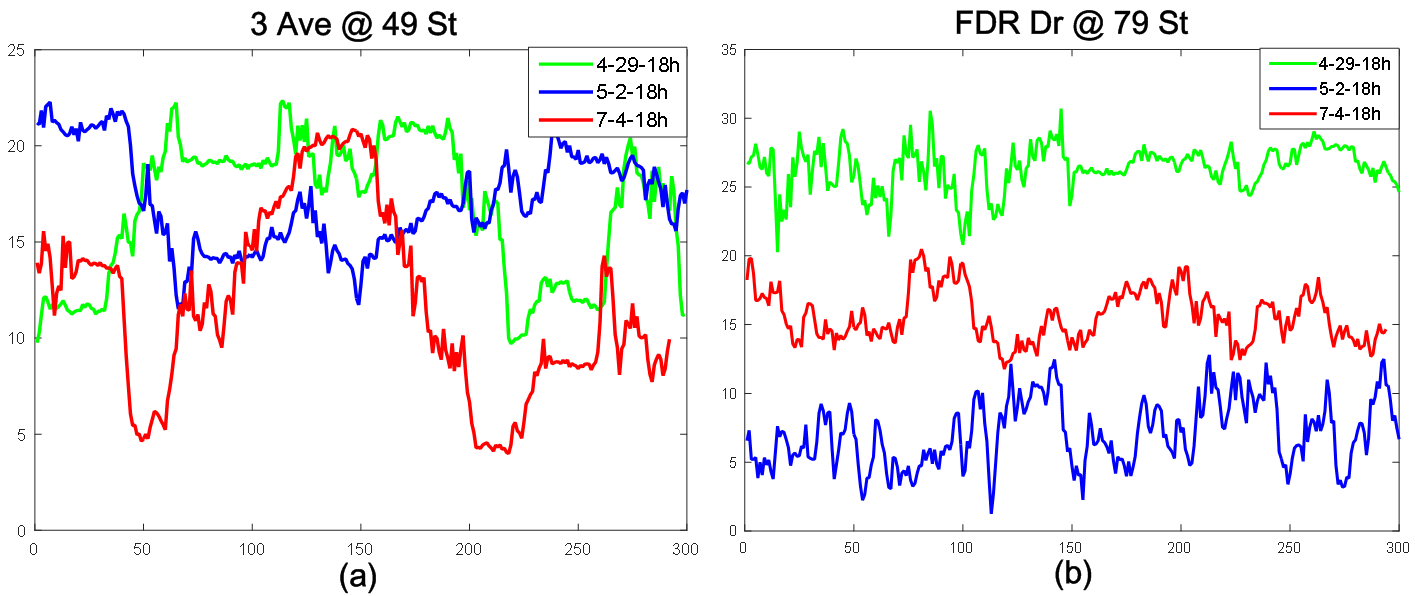}
\end{center}
\vspace{-0.3cm}
\caption{Independence Day traffic density detection. X-axis: frame index. Y-axis: vehicle count.}
\vspace{0.1cm}
\label{fig:7-4}
\end{figure}

\textbf{Special Event Detection}.
An interesting application of traffic density estimation is to detect the change of traffic density when special event happens in the city. To verify the capability to detect such changes of our method, we test FCN-MT on two cameras and multiple days. From the results we find that traffic density on July 4th 18h is different from that on other normal days, as illustrated in Figure \ref{fig:7-4}. For the camera in downtown (3Ave@49st), the traffic density on July 4th is averagely lower than that of other days, and the traffic can be very sparse periodically. This corresponds to the fact that several roads around 3Ave@49st are closed after 3 pm due to fireworks show on Independence Day, resulting in less traffic around. For the camera on parkway (FDR Dr @ 79st), the average traffic is less then Friday (4-29), but more than a normal Monday (5-2). As July 4th is also Monday, it is supposed to have similar traffic as May 2nd. The detected increase of the traffic density on July 4th corresponds to the fact that FDR under 68 St is closed, resulting in more traffic congested to FDR above 68 St. All those observations verify that our method can detect traffic density change when special events happen.

\setlength{\textfloatsep} {0pt plus 2pt minus 5pt}

\subsection{Quantitative Evaluations on TRANCOS}
To verify the efficacy of our methods, we also evaluate and compare the proposed methods with baselines on a public dataset, TRANCOS\cite{onoro2016towards}. TRANCOS provides a collection of 1244 images of different traffic scenes, obtained from real video surveillance cameras, with a total of 46796 annotated vehicles. The objects have been manually annotated using dots. It also provides a region of interest (ROI) per image, defining the region considered for the evaluation. This database provides images from very different scenarios and no perspective maps are provided. The ground truth object density maps are generated by placing a Gaussian Kernel in the center of each annotated object\cite{guerrero2015extremely}.

\begin{figure}[t]
\setlength{\abovecaptionskip}{0.cm}
\setlength{\belowcaptionskip}{-0.5cm}
\begin{center}
\includegraphics[width=0.95\columnwidth, height = 6.5cm]{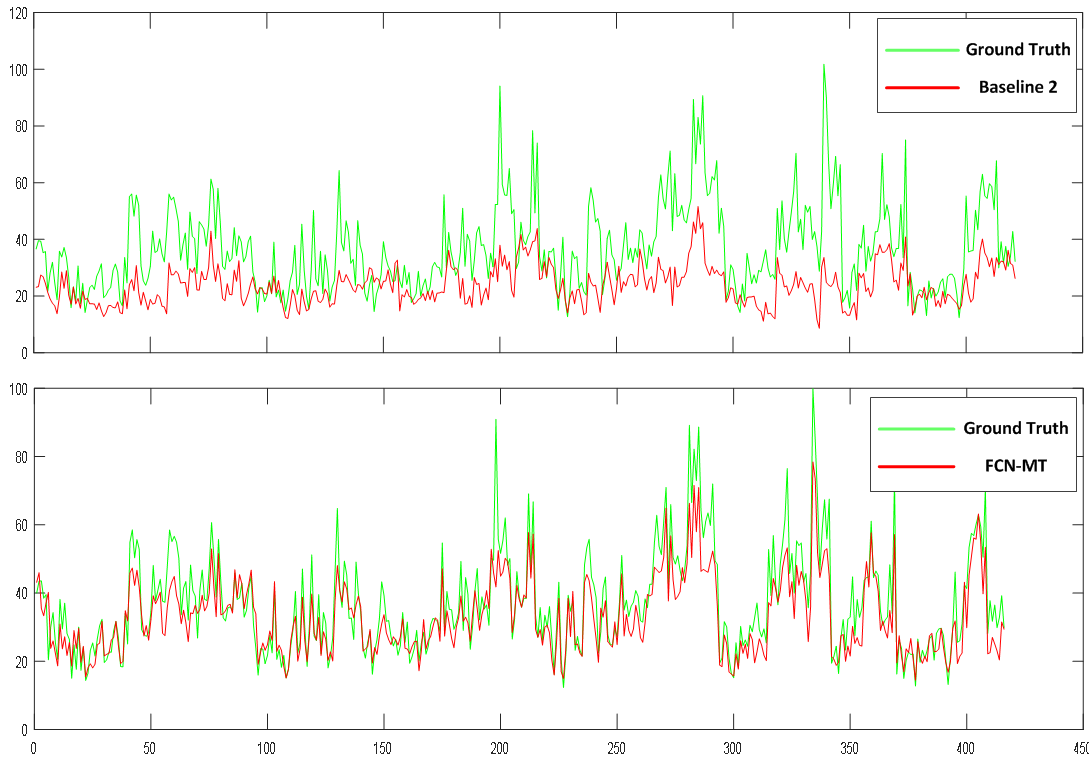}
\end{center}
\vspace{-0.3cm}
\caption{Comparing FCN-MT and Baseline 2-Hydra. X-axis: frame index. Y-axis: vehicle count.}
\label{fig:TranFCN}
\end{figure}

We compare our methods with baselines in Table \ref{tb:tb5}. Baseline 1 and OPT-RC have the same settings as evaluated in WebCamT. Baseline 2-CCNN is a basic version of the network in \cite{onoro2016towards}, and Baseline 2-Hydra augments the performance by learning a multi-scale non-linear regression model. FCN-ST is the single task implementation (only vehicle density estimation) of FCN-MT for ablation analysis. Baseline 2-CCNN, Baseline 2-Hydra, FCN-ST, and FCN-MT are trained on 823 images and tested on 421 frames following the separation in \cite{onoro2016towards}. From the results, we see that FCN-MT significantly reduces the MAE from 10.99 to 5.31 compared with Baseline 2-Hydra. FCN-MT also outperforms the single task method FCN-ST and verifies the efficacy of multi-task learning. From Figure \ref{fig:TranFCN} we can also see that the estimated counts from FCN-MT fit the ground truth better than the estimated counts from Baseline 2. OPT-RC outperforms Baseline 1 and obtains results comparative to Baseline 2.

\begin{table}[]
\centering
\small
\caption{Results comparison on TRANCOS dataset}
\label{tb:tb5}
\begin{tabular}{|l|l|l|}
\hline
Method & MAE & ARA \\ \hline
Baseline 1 & 13.76 & 0.6412 \\ \hline
Baseline 2-CCNN & 12.49 & 0.6743 \\ \hline
Baseline 2-Hydra & 10.99 & 0.7129 \\ \hline
OPT-RC &12.41  & 0.6674 \\ \hline
FCN-ST & 5.47 & 0.827\\ \hline
FCN-MT & 5.31 & 0.856\\ \hline
\end{tabular}%
\end{table}

\setlength{\textfloatsep} {0pt plus 20pt minus 5pt}

\subsection{Quantitative Evaluations on UCSD Dataset}
To verify the generalization of the proposed methods on different counting tasks, we evaluate the proposed methods on crowd counting dataset UCSD\cite{chan2008privacy}. This dataset contains 2000 frames chosen from one surveillance camera, with frame size $158 \times 238$ and frame rate $10 fps$. Average number of people in each frame is around $25$. 

By following the same settings with \cite{chan2008privacy}, we use frames from 601 to 1400 as training data, and the remaining 1200 frames as test data. OPT-RC has the same setting as evaluated in WebCamT. ROI mask is used on the input images and combined feature maps in FCN-MT and OPT-RC. Table \ref{tb:tb6} shows the results of our methods and existing methods. From the results we can see OPT-RC outperforms the non-deep learning based methods, but is less accurate than the deep learning-based methods in\cite{zhang2015cross}. FCN-MT outperforms all the non-deep learning based methods and gets comparative accuracy with the deep learning-based methods in\cite{zhang2015cross}. These results show that though our methods are developed to solve the challenges in webcam video data, they are also robust to other type of counting tasks.

\begin{table}[]
\centering
\small
\caption{Results comparison on UCSD dataset}
\label{tb:tb6}
\begin{tabular}{|l|l|l|}
\hline
Method & MAE & MSE \\ \hline
Kernel Ridge Regression \cite{an2007face} & 2.16 & 7.45 \\ \hline
Ridge Regression \cite{chen2012feature} & 2.25 & 7.82 \\ \hline
Gaussian Process Regression \cite{chan2008privacy} & 2.24 & 7.97 \\ \hline
Cumulative Attribute Regression \cite{chen2013cumulative} & 2.07 & 6.86 \\ \hline
Cross-scene DNN\cite{zhang2015cross} & 1.6 & 3.31 \\ \hline
OPT-RC & 2.03 & 5.97 \\ \hline
FCN-MT & 1.67 & 3.41 \\ \hline
\end{tabular}
\end{table}
\vspace{0.4cm}

\vspace{-0.2cm}
\section{Comparison of OPT-RC and FCN-MT}

From the extensive experiments, we highlight some difference between OPT-RC and FCN-MT:
(i) OPT-RC can learn the geometry information by learning different weight for different blocks in the image.
(ii) As the handcrafted SIFT feature is not discriminative enough to distinguish background and foreground, OPT-RC heavily relies on background subtraction. Nevertheless FCN-MT extracts hierarchical feature maps. The combined and re-weighted features are quite discriminative for foreground and background. Thus FCT-MT does not require background subtraction.
(iii) The learned density map of FCN-MT suffers from checkerboard artifacts in some cases.

Despite of these differences, FCN-MT and OPT-RC still have strong connection: both methods map the image into vehicle density map and overcome the challenges of webcam video data. FCN-MT replaces the BG subtraction, feature extractor, and block-regressors of OPT-RC with fully convolutional networks. Both methods avoid detecting or tracking individual vehicle, and adapt to different vehicle scales. For the future research, domain transfer learning will be explored to enable the model more robust to multiple cameras.

% The last convolution 1x1 layer of FCN-MT can be understood as rank one regression, mapping the final feature map into density map with the same weight for all the locations, while OPT-RC relaxes the rank one constraint and learns different weights for different locations. 

% As show in Figure \ref{fig:fig17},

\iffalse
\begin{figure}[t]
\setlength{\abovecaptionskip}{0.cm}
\setlength{\belowcaptionskip}{-0.5cm}
\begin{center}
\includegraphics[width=0.8\columnwidth, height = 3cm]{figs/Comparison.png}
\end{center}
\vspace{-0.3cm}
\caption{Combination of FCN-MT and OPT-RC.}
\label{fig:fig17}
\end{figure}
\setlength{\textfloatsep} {0pt plus 2pt minus 15pt}
\fi

\noindent\footnotesize{\textbf{Acknowledgment} This research was supported in part by Funda\c{c}\~{a}o para a Ci\^{e}ncia e a Tecnologia (project FCT [SFRH/BD/113729/2015]). Jo\~{a}o P. Costeira is partially supported by project [Lx-01-0247-FEDER-017906] SmartCitySense funded by ANI.}

{\small
\bibliographystyle{ieee}
%\bibliography{egbib.bib}
\bibliography{CVPR camera ready2.bbl}
}

\end{document}